\documentclass[final]{cvpr}

\usepackage{times}
\usepackage{epsfig}
\usepackage{graphicx}
\usepackage{amsmath}
\usepackage{amssymb}
\usepackage{mathtools}
\usepackage{pifont}
\usepackage{amssymb}
\usepackage{booktabs}
\usepackage{caption}
\newcommand\mat[1]{\mathbf{#1}}

\newcommand{\good}{\checkmark}
\newcommand{\bad}{\ding{55}}
\usepackage{subfig}
\usepackage[pagebackref=true,breaklinks=true,colorlinks=true,bookmarks=false,urlcolor=black]{hyperref}
\def\cvprPaperID{21} % *** Enter the CVPR Paper ID here
\def\confYear{CVPR 2021}
\begin{document}

%%%%%%%%% TITLE
\title{In-Hindsight Quantization Range Estimation for Quantized Training}

\author{Marios Fournarakis, Markus Nagel \\
Qualcomm AI Research\thanks{\scriptsize Qualcomm AI Research is an initiative of Qualcomm Technologies, Inc.}\\
{\tt\small mfournar@qti.qualcomm.com, markusn@qti.qualcomm.com}
}

\maketitle

%%%%%%%%% ABSTRACT
\begin{abstract}
   Quantization techniques applied to the inference of deep neural networks have enabled fast and efficient execution on resource-constraint devices. The success of quantization during inference has motivated the academic community to explore fully quantized training, i.e. quantizing back-propagation as well. However, effective gradient quantization is still an open problem. Gradients are unbounded and their distribution changes significantly during training, which leads to the need for dynamic quantization. As we show, dynamic quantization can lead to significant memory overhead and additional data traffic slowing down training. We propose a simple alternative to dynamic quantization, in-hindsight range estimation, that uses the quantization ranges estimated on previous iterations to quantize the present. Our approach enables fast static quantization of gradients and activations while requiring only minimal hardware support from the neural network accelerator to keep track of output statistics in an online fashion.
   It is intended as a drop-in replacement for estimating quantization ranges and can be used in conjunction with other advances in quantized training. We compare our method to existing methods for range estimation from the quantized training literature and demonstrate its effectiveness with a range of architectures, including MobileNetV2, on image classification benchmarks (Tiny ImageNet \& ImageNet).

\end{abstract}

%%%%%%%%% BODY TEXT
\section{Introduction}
\label{sec:intoduction}
Deep Neural Networks (DNNs) have become the state-of-the technique for a wide range of Computer Vision (CV) applications, such as image recognition, object detection or machine translation. However, as the accuracy and effectiveness of these networks grow, so does their size. The high computational cost and memory footprint can impede the deployment of such networks to resource-constrained devices, such as smartphones, wearables or drones. Fortunately, in recent years low-bit network quantization for neural network inference has been extensively studied and in combination with dedicated hardware utilizing efficient fixed-point operations, they have succeeded in accelerating DNN inference \cite{Jacob,MSFT,adaround,dfq,lsq}.

However, training DNNs still predominately relies on  floating-point format. As we move towards a world of privacy-preserving and personalized AI, we can expect increased requirements for training on edge devices that do not have the computational resources of servers. This raises the need for more power-efficient training techniques for neural networks. Quantizing back-propagation can provide considerable acceleration and power efficiency but the noise induced by gradient quantization can be detrimental to the network's accuracy \cite{zhu2019unified, FX_backprop_trainig}. Recent work has shown that \textit{quantized training}, can achieve accuracy within $1\%-2\%$ of floating-point (FP32) training in a range of  CV tasks and models \cite{zhu2019unified, IBM_custom_FP, chen2020statistical}. In most cases, this is possible by quantizing the weights, activations and activation gradients to 8-bits and maintaining certain operations in floating-point, such as batch-normalization \cite{batch-norm} or weight updates. 

A key challenge present throughout the quantized training literature is how to set the quantization range for gradients \cite{chen2020statistical, zhu2019unified, sakr2018pertensor, banner2018scalable, WAGE, WAGEUBN}. Because gradients are unbounded, choosing the quantization range appropriately is important to keep the quantization error in check. Some existing methods use the min-max range of the gradient tensor \cite{WAGE,WAGEUBN, zhou2018dorefanet} whereas others use a moving average of the tensor's statistics \cite{FX_backprop_trainig}. \cite{chen2020statistical} goes even further and propose a per-sample quantization of gradient tensor. However, in all cases there is a common theme: to determine the quantization parameters of the tensor, these methods require access to the unquantized gradient tensor. In other words, they perform \textit{dynamic quantization}. 

Dynamic quantization can reduce the quantization error as the quantization grid is better utilized but comes with significant memory overhead \cite{QuantSurvey2021,pytorch_quant,tensorflow}: the quantization range depends on the full tensor output, therefore, the entire full precision tensor needs to be written to memory before it can be quantized. For typical layers in common DNNs, this can lead up to $8 \times$ more memory transfer.
In section \ref{sec:dynamic_quantization}, we discuss in more detail the implications of dynamic quantization for fixed-point accelerator hardware and quantify the associated overhead.

In this work, we provide a hardware-friendly alternative to dynamic quantization for quantized training, called \textit{in-hindsight range estimation} that better utilizes the efficiencies provided by modern fixed-point accelerators. Our approach uses the quantization ranges from previous iterations to quantize the current tensor. This method allows us to use pre-computed quantization ranges to accelerate training and reduce the memory overhead. We use a moving average of the quantization range and update the ranges with statistics (min and max) extracted from the accumulator in an online fashion. These statistics need to be extracted from the accumulator before the quantization step, which may require appropriate hardware support.

We evaluate our proposed framework on Tiny ImageNet and ImageNet datasets. We show that our method achieves comparable accuracy to dynamic quantization when applied to activations and gradients. Our approach is intended as a drop-in replacement for estimating quantization ranges and can be used in conjunction with other advances in quantized training.

\section{Related Work}
\label{sec:related_work}
In this section, we outline some of the relevant work in the area of quantized training. We split the contribution into two sections. First, we discuss general advances in quantized training and, second, we focus on the details of quantization range estimation by relevant work. We concentrate on activation and gradient quantization because they rely on dynamic quantization. Weights can be quantized offline as they do not depend on the input data. 

\subsection{Quantized Training}
One of the earliest attempts in fully quantized training dates back to 2015 \cite{gupta2015deep}. The authors train a  3-layer convolutional network on CIFAR-10 and MNIST and introduce stochastic rounding as an unbiased quantization operator.  
% Back in 2015 \cite{gupta2015deep} first perform fully quantized training of a  3-layer convolutional on CIFAR-10 and MNIST and introduce stochastic rounding as unbiased quantization operator. 
DoReFa-Net \cite{zhou2018dorefanet} train low-bit AlexNet with BatchNorm on ImageNet but struggle to close the gap to full precision models when training from scratch. Range Normalization is introduced by \cite{banner2018scalable} as a fixed-point friendly alternative to BatchNorm and they observe no accuracy drop when training ResNet-50 on ImageNet. \cite{sakr2018pertensor} formulate a theoretical framework for finding the optimal bit-width for all quantized tensor in fully quantized training and perform the weight update in fixed-point but keep BatchNorm operations in floating-point. WAGE \cite{WAGE} adopt a layer-wise scaling factor instead of using BatchNorm and quantize gradients to 8-bits while keeping the weight update to 16-bits. WAGEUBN \cite{WAGEUBN} expand on WAGE and implement a quantized implementation of BatchNorm for the forward and backward pass. Deviation Counteractive Learning Rate Scaling \cite{zhu2019unified} uses an exponential decaying learning rate based on the cosine distance between the full precision and quantized gradients to stabilize training. The authors measure a $22\%$ training time reduction for ResNet-50 on ImageNet and apply their training framework to MobileNetV2 and Inception V3. Recently, it was shown that ResNet-50 can be trained within $1\%$ of FP32, using a hybrid of INT4 forward quantization and a novel Radix-4 FP4 format for the gradients \cite{IBM_custom_FP}.

\subsection{Quantization Range Estimation}
In earlier attempts \cite{gupta2015deep}, a fixed-point format with a fixed decimal point was used for activations and gradients throughout training.
% \cite{gupta2015deep} use a fixed-point format with a fixed decimal point for activations and gradients throughout training. 
DoReFa-Net clip the activations to the $[0,1]$ range and use the dynamic min-max range to quantize the gradients. WAGE and WAGEUBN use a pre-defined scale factor $\alpha$ for each layer's activation which depends on the network's structure and the min-max range for the gradients. \cite{sakr2018pertensor} fix the activation range to $[0,2]$ and use an exponential moving average of the gradient's standard deviation to calculate the gradient bit-width. Models with RagenNorm \cite{banner2018scalable} split the activation and gradient tensor in chunks and use the average minimum and maximum of the chunks to estimate the quantization range. \cite{FX_backprop_trainig} use an exponential moving average of the maximum absolute value for activation and gradients. Direction sensitive gradient clipping \cite{zhu2019unified} periodically updates the gradient clipping range by searching for the clipping values that minimizes the angle between the FP32 and quantized gradients. To the best of our knowledge, this is the only existing method that to an extend avoids dynamic quantization

In all previous cases, the statistics are computed over the complete tensor. \cite{chen2020statistical} observe that per-tensor quantization of gradients does not utilize the quantization grid efficiently and instead propose a per-sample quantization and a Block Householder decomposition of the gradient tensor to better spread the signal across the tensor. 

\begin{figure*}
\begin{center}
\includegraphics[width=0.7\linewidth]{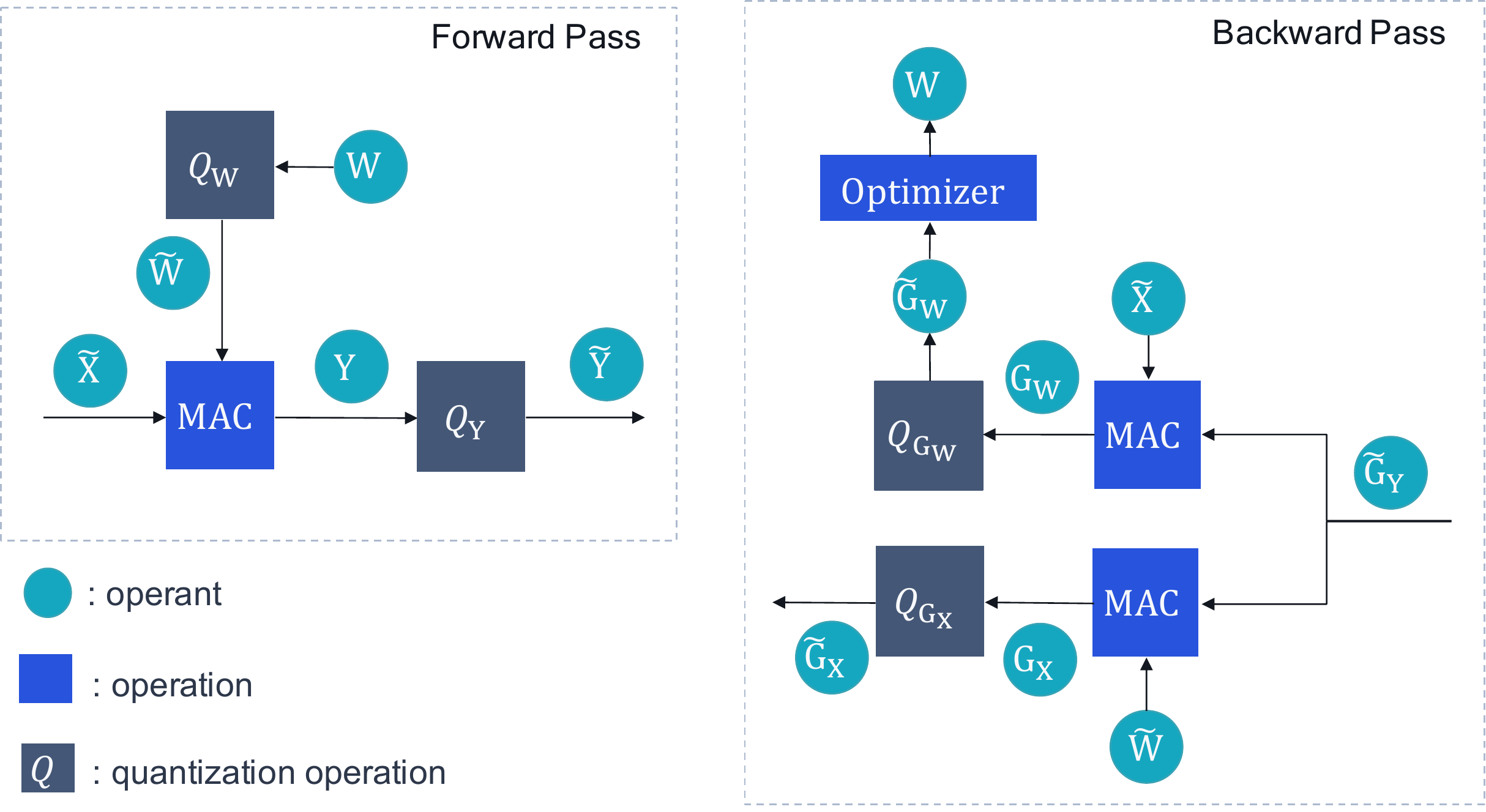}
\end{center}\vspace{-.4cm}
   \caption{Forward and backward pass for quantized training pipeline.}
\label{fig:qt_framework}
\end{figure*}

\section{Problem Formulation}
\label{sec:problem_formulation}
In this section, we outline how quantized training works using typical fixed-point accelerator architecture and explain how dynamic quantization can lead to memory and compute overhead. In this work, we concentrate on hardware dedicated to fixed-point operations to extract the biggest power gains from quantized tensor operations. 

\subsection{Quantized Training Framework}
\label{sec:quant_training_framework}
In figure \ref{fig:qt_framework} we present a compute graph for the forward (left) and backward pass (right) for our quantized training framework. The forward pass is very similar to that quantization-aware training (QAT) \cite{krishnamoorthi,Jacob}. The weights $\mat{W}$ are typically stored in higher precision (16-bits or FP32) to allow the accumulation of small gradients during training. The weight tensor is quantized to a low-bit representation $\widetilde{\mat{W}}$ through the quantization function $Q_{\mat{W}}(\cdot)$ before it is loaded into the Multiply \& Accumulate (MAC) array.
The quantized input $\widetilde{\mat{X}}$ is also loaded into the MAC array to compute the linear operations of the layer. As $\widetilde{\mat{X}}$ and $\widetilde{\mat{W}}$ are typically larger than the MAC array, the output is calculated over multiple compute cycles. The output of the MAC array  $\mat{Y}$ is typically in 32-bits and is thus followed by another quantization step, $Q_{\mat{Y}}(\cdot)$, to convert it to the required bit-width.

During the backward pass, the quantized activation gradient $\widetilde{\mat{G}}_{\mat{Y}}$ is used to calculated the weight gradient $\mat{G}_{\mat{W}}$ and input gradient $\mat{G}_{\mat{X}}$. The input gradient is typically quite large and is always quantized to a low-bit representation $\widetilde{\mat{G}}_{\mat{X}}$ before it is propagated to the preceding layer. The weight gradient can be re-quantized to a lower-bit representation, but it is also common in literature to keep it in full precision. In this work, we also keep the weight gradient in FP32 and we denote $Q_{\mat{G}_{\mat{X}}}(\cdot)= Q_{\mat{G}}(\cdot)$ for clarity. 

For fully quantized training, we need to specify the quantization ranges of at least three quantizers $Q_{\mat{W}}(\cdot)$, $Q_{\mat{Y}}(\cdot)$ and $Q_{\mat{G}}(\cdot)$.  Because the weights are independent of the data, the quantization range for the weights can be pre-computed and be stored in memory. However, this is not the case for the activations and gradient, as they depend on the current batch. To address this issue most existing techniques assume \textit{dynamic quantization} for these quantizers. In the following section, we discuss what exactly is dynamic quantization and its implications for quantized training. It is important to be able to adjust the quantization ranges of gradients during training because the gradient distribution changes significantly during training \cite{IBM_custom_FP, zhu2019unified, FX_backprop_trainig}.

\subsection{Dynamic Quantization}
\label{sec:dynamic_quantization}
\begin{figure}[t]
\begin{center}
  \includegraphics[width=0.6\linewidth]{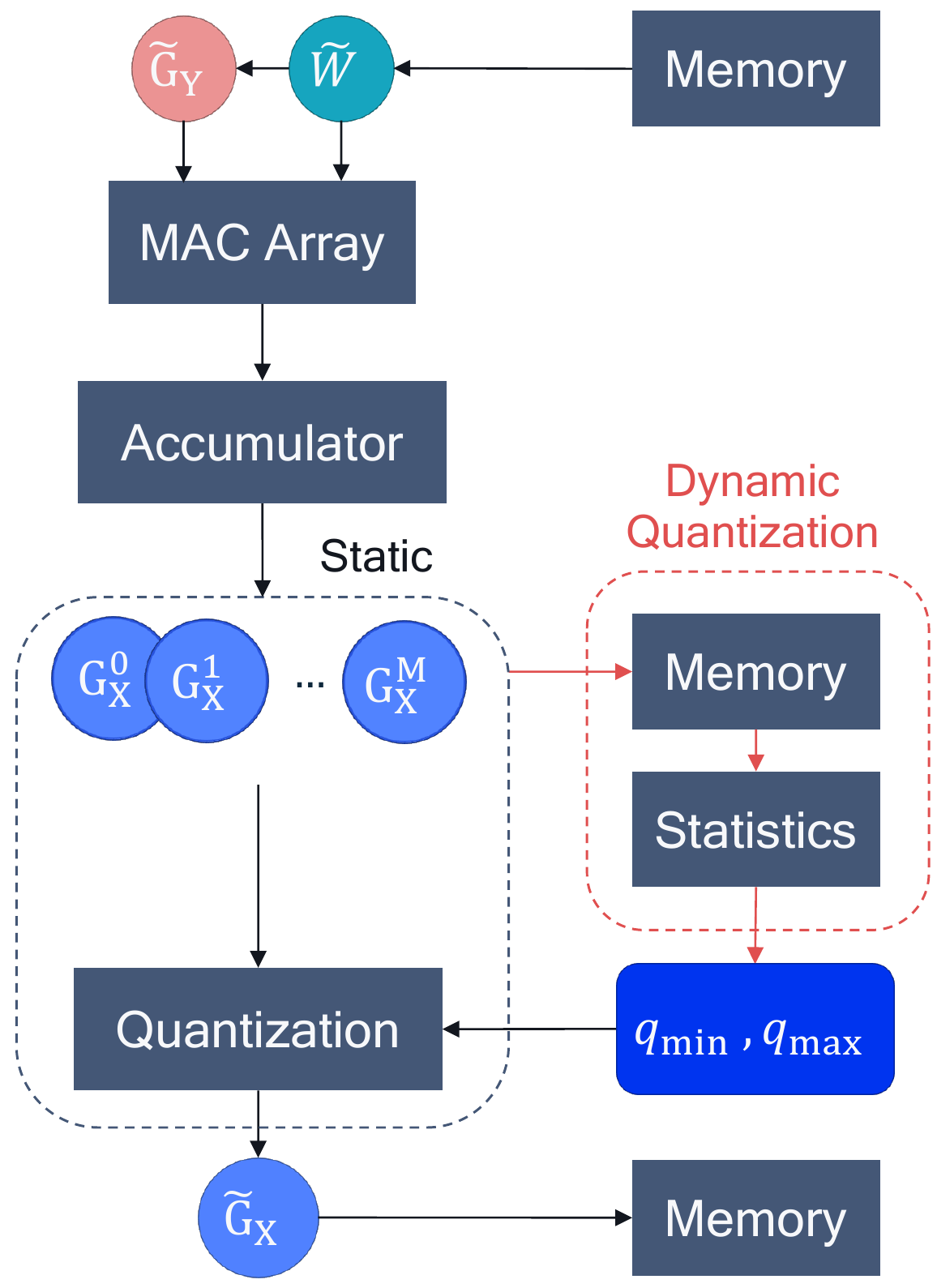}
\end{center}\vspace{-.4cm}
  \caption{Quantized neural network accelerator diagram. The MAC array size is fixed which means that output tensor can only be computed in slices. In the case of static quantization the quantization parameters are know in advance and the accumulator output is directly quantized. For dynamic quantization all outputs have to be written to memory before they can be quantized.}
\label{fig:dynamic_quant}
\end{figure}

Dynamic quantization uses the statistics of the full precision tensor to quantize it. Assuming a quantization range of $(q_{\text{min}},q_{\text{max}})$, then in its simplest form dynamic quantization uses the \textit{min-max range} of the full tensor $\mat{G}$:
\begin{align}
\label{eq:current_min_max}
    q_{\text{min}}= \min{\mat{G}}, \quad q_{\text{max}}= \max{\mat{G}}
\end{align}
Figure \ref{fig:dynamic_quant} illustrates how matrix multiplication is computed in a typical fixed-point accelerator. The logic consists of a fixed-size MAC array and accumulators. Typically, in deep learning the size of matrices that are multiplied exceed the size of the MAC array. For this reason, the computation takes places in slices until the whole matrix multiplication is completed. The output of the accumulator is typically in higher bit-width (32-bits) to avoid overflow and is normally followed by a quantization step. 

In the case of \textit{static} quantization, the quantization ranges of the output $(q_{\text{min}},q_{\text{max}})$ are known in advance. Therefore, each output from the accumulator can be quantized directly and be stored in memory in a low-bit representation. 
On the other hand, in \textit{dynamic} quantization, the ranges depend on the output itself. To extract the necessary statistics, all the outputs of the accumulator have to first be written in memory. After the quantization ranges have been calculated, the tensor is brought back to the compute unit to be quantized and then stored back in memory. 

It is evident, that dynamic quantization can lead to significant memory overhead and extra data movement. Our analysis in section \ref{sec:memory_transfer_comparios} shows that dynamic quantization can lead up to $8 \times$ additional memory transfer depending on the size of the layer. A study between static and dynamic quantization for an MLP in PyTorch 1.4 on a CPU also showed that dynamic quantization leads on average to a $20\%$ latency increase for inference.  Unfortunately,  dynamic quantization for convolutions is not implemented in PyTorch, which might be due to the even larger overhead associated with large feature maps.

%%%% TO DO:
%  Iclud metrics for additional data movement.

\section{In-Hindsight Range Estimation}
\label{in_hindsight_range_estimation}
\begin{figure}
\begin{center}
  \includegraphics[width=0.6\linewidth]{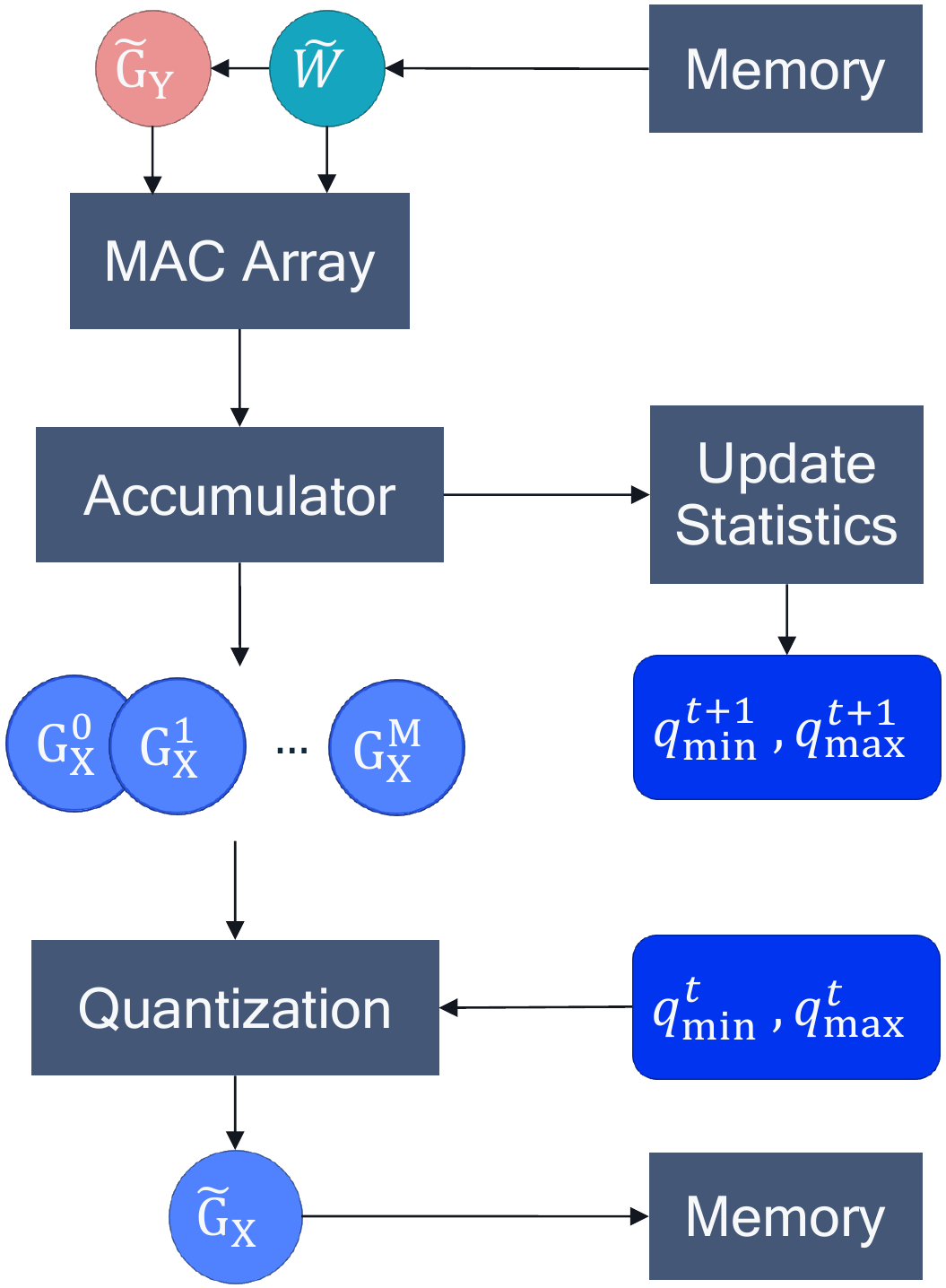}
\end{center}\vspace{-.4cm}
  \caption{General framework for in-hindsight range estimation. Using pre-computed quantization ranges enables efficient static quantization of the output. Additional logic is needed at the accumulator level update the ranges for the next step.}
\label{fig:in_hindsight_graphic}
\end{figure}

Our proposed method aims at preventing the need for dynamic quantization during quantized training and unlock the speed-ups provided by dedicated fixed-point hardware. The method involves two key steps:
\begin{enumerate}
    \item Use \textit{pre-computed} quantization parameters to quantize the current tensor.
    \item Extract \textit{statistics} from the current tensor in an  online fashion to update  the quantization parameters for the next iteration.
\end{enumerate}
Figure \ref{fig:in_hindsight_graphic} shows a general framework of how in-hindsight range estimation can be implemented in hardware. The benefit of this approach is that the pre-computed quantization enables fast and efficient static quantization. The required statistics should be easy to calculate at the accumulator or quantization level, to reduce the computational overhead of the method. Such statistics can be the min and max statistics or the saturation ratio\footnote{The proportion of values that lie outside the quantization grid.}. In some cases, extracting these statistics may require appropriate hardware logic around the accelerator.

% \subsection{Saturation Based}
% One of the simplest statistics that can be extracted is the \textit{saturation ratio} $\alpha$, which is the proportion of values that are clipped during the quantization operation:
% \begin{equation}
%  \label{eq:saturation_ratio}
%     \alpha = \frac{ |\mat{G}<q_{\text{min}} | + | \mat{G}>q_{\text{max}}|}{| \mat{G} |}
% \end{equation}
% where $|\cdot|$ represents the number of elements in the tensor. If the saturation ratio lies within a acceptable range $[\alpha_{\text{min}},\alpha_{\text{max}} ]$, the quantization ranges need not be updated. If these limits are exceeded consistently (for more than $\tau$ iterations), the we update the quantization ranges for the next iteration. To update the quantization ranges in a way non-dynamic way we use the inverse cumulative density function ($\text{CDF}^{-1}$) of a known distribution. Based on our analysis of activation gradient histograms  we choose the CDF of the Laplace distribution. Our observation is also supported by a statistical analysis of gradients from \cite{zhu2019unified}. Our method is summarised in the following equations:
% \begin{equation}
%     \label{eq:in_hidnsight_saturation}
% \end{equation}

\subsection{In-Hindsight Min-Max}
\label{sec:past_min_max}
We propose an instance of our framework that uses the min-max statistics, which we call \textit{in-hindsight min-max}. In this method, we define the quantization range as the exponential moving average of the tensor's min-max statistics. To quantize the tensor at step $t$, we use the estimate of quantization ranges from the previous iteration. While the output is computed,  appropriate logic keeps track of the min-max statistics from the accumulator. These statistics are then used to update the quantization ranges for the next iteration as soon as the complete output tensor has been calculated. The quantization ranges are calculated as:
\begin{align}
   q^t_{\text{min}} &= (1-\eta) \min{\mat{G}^{t-1}} + \eta q^{t-1}_{\text{min}}\\
   q^t_{\text{max}} &= (1-\eta) \max{\mat{G}^{t-1}} + \eta q^{t-1}_{\text{max}}
    \label{eq:past_min_max}
\end{align}
where $\eta$ is the momentum term. To initialize the scheme ($t=0$), we can use the min-max range of the first batch, namely $q^0_{\text{min}} = \min{\mat{G}^0}$ and $q^0_{\text{max}} = \max{\mat{G}^0}$.  If we replace the $\mat{G}^{t-1}$ with $\mat{G}^{t}$ in the above equations we end up with a dynamic quantization method called \textit{running min-max}, which is common in post-training quantization \cite{krishnamoorthi} and  adopted for gradient quantization by \cite{FX_backprop_trainig}. 

\section{Experiments}
\label{sec:experiments}
To verify the effectiveness of our method we conduct experiments on the Tiny ImageNet \cite{TinyImageNet} and full ImageNet \cite{imagenet} benchmarks. We train ResNet18 on ImageNet and a modified version on Tiny ImageNet \cite{TinyResNet18}. We also train our own version VGG16 \cite{vgg} and MobileNetV2 \cite{MobileNetV2} on Tiny ImageNet.

\subsection{Range Estimation Methods Comparison}
\label{sec:methods_comparison}
Typically, in literature, we only observe final results for the fully quantized setting,  making it difficult to assess the impact of the individual quantization choices for each tensor. In this section, we aim at better understanding the impact of the individual range estimation methods for either gradients or activations quantization at the final accuracy. We compare our hardware friendly method, in-hindsight min-max range estimation, to the commonly used dynamic quantization methods: current min-max \cite{zhou2018dorefanet, WAGE,WAGEUBN, zhu2019unified}  and running min-max \cite{krishnamoorthi, FX_backprop_trainig} estimators. 

We further compare to Direction Sensitive Gradient Clipping (DSGC) \cite{zhu2019unified}. DSGC searches for the optimal clipping values that maximizes the cosine similarity between the quantized and full precision tensor. In our implementation of the method, we use golden section search to find the optimal quantization ranges, as the authors do not provide implementation details.  Because of the computational cost of such optimization, the quantization range is only updated periodically. We use an update interval of $100$ iteration as per the authors. Note, this method is a hybrid between static quantization and dynamic quantization. It uses HW friendly static quantization for most iterations but the update step can be very expensive, as it requires estimating the objective function (cosine similarity) at multiple clipping thresholds.

We also experimented with using an exponential moving average of the gradient variance \cite{sakr2018pertensor}  to define the quantization ranges. However, we found that it made training unstable despite an extensive hyper-parameter search for momentum and the number of standard deviations.

\paragraph{Experimental Setup}  
We conduct our study on Tiny ImageNet with the modified ResNet18 and train for $90$ epochs using SGD with initial learning rate of 0.1 and momentum of 0.9. The learning rate is reduced by a factor of $10$ at epochs $30$ and $60$ and we use a weight decay of $1\text{e-}4$. For all different range estimation methods on the gradient and activations tensor, we tune the hyper-parameter (e.g. momentum term) individually and present the  results with the best hyper-parameters. 

\paragraph{Gradient Quantization}
To see the effect of range estimation for gradient quantization, we keep the forward pass in full precision and only quantize the activation gradient to 8-bit using asymmetric uniform quantization with stochastic rounding \cite{gupta2015deep}.

We see that in hindsight min-max performs on par and even better than the other dynamic quantization methods. Its performance is also comparable to DSGC, which is a hybrid between dynamic and static quantization method. However, in-hindsight range estimation relies on simple statistics, whereas DSGC requires a computationally intensive parameter search. We also found that under certain seeds training with DSGC can become quite unstable, which is also reflected by the larger standard deviation of the final result. Current min-max underperforms all other methods. This analysis demonstrates that switching to a better range estimator could be very beneficial  before attempting more complex solutions.
 
It is also interesting to observe that gradient quantization leads to accuracy improvements compared to FP32 training across the board likely due to its regularization effect.

\begin{table}
\begin{center}
\begin{tabular}{lcc}
\toprule
Method  & Static & Val. Acc. (\%) \\
\midrule
FP32 & n.a. & 58.97 $\pm$ 0.13 \\
Current min-max  & \bad & 59.14 $\pm$ 0.23\\
Running min-max  & \bad  & 59.25 $\pm$ 0.55 \\
DSGC \cite{zhu2019unified} & \bad & 59.35 $\pm$ 0.95\\
\midrule
In-hindsight min-max & \good & 59.46 $\pm$ 0.71 \\
\bottomrule 
\end{tabular}
\end{center}\vspace{-.4cm}
\caption{Gradient quantization range estimators comparison. Validation accuracy (average of 5 seeds) and standard deviation for ResNet-18 on Tiny ImageNet. }
\end{table}

\begin{table}
\begin{center}
\begin{tabular}{lcc}
\toprule
Method  & Static & Val. Acc. (\%) \\
\midrule
FP32 & n.a. & 58.97  $\pm$ 0.13\\
Current min-max  & \bad & 59.00 $\pm$ 0.31\\
Running min-max  & \bad &59.28  $\pm$ 0.25\\
DSGC \cite{zhu2019unified}  &  \bad & 59.09 $\pm$ 0.01\\ 
\midrule
In-hindsight min-max & \good & 59.30  $\pm$ 0.19 \\
\bottomrule
\end{tabular}
\end{center}\vspace{-.4cm}
\caption{Activation quantization range estimator comparison. Validation accuracy (average of 5 seeds) and standard deviation for ResNet-18 on Tiny ImageNet.}
\label{tab:tiny_imagenet}
\end{table}

\begin{table*}[t]
\begin{center}
\begin{tabular}{llcccc}
\toprule
Gradient Method & Activation Method & Static & ResNet18 & VGG16 & MobileNetV2 \\
\midrule
FP32 & FP32 & n.a.  & 58.97 $\pm$ 0.13   & 53.79 $\pm$ 0.30  & 59.61 $\pm$ 0.37\\
Current min-max  & Current min-max &  \bad & 58.77 $\pm$ 0.73   & 53.28 $\pm$ 0.43  & 58.88 $\pm$ 0.73 \\ 
Running min-max & Running min-max & \bad &59.20 $\pm$ 0.25   & 53.36 $\pm$ 0.27  &  59.69 $\pm$ 0.09 \\
DSGC \cite{zhu2019unified} & Current min-max & \bad &59.07 $\pm$ 0.33  & 52.84 $\pm$ 0.28 & 59.10 $\pm$ 0.44  \\
\midrule
In-hindsight min-max  &  In-hindsight min-max & \good &58.99 $\pm$ 0.44   & 53.25 $\pm$ 0.41 &  59.28 $\pm$ 0.20 \\
\bottomrule
\end{tabular}
\end{center}\vspace{-.4cm}
\caption{Results on Tiny ImageNet. Average validation accuracy (\%) with standard deviation: 5 seeds for ResNet18 \& VGG16 and 3 seeds for MobilenetV2.}
\label{tab:imagenet}
\end{table*}

\begin{table*}
\begin{center}
\begin{tabular}{llcc}
\toprule
Gradient Method & Activation Method & Static & ResNet18  \\
\midrule
FP32 & FP32 & n.a. &69.75 \\
Current min-max  & Current min-max & \bad & 69.21 $\pm$ 0.06  \\
Running min-max & Running min-max & \bad & 69.35 $\pm$ 0.16 \\
\midrule
In-hindsight min-max  & In-hindsight min-max & \good & 69.37 $\pm$ 0.11 \\
\bottomrule
\end{tabular}
\end{center}\vspace{-.4cm}
\caption{Validation accuracy (\%) results on  ImageNet. Average validation accuracy (\%) of 3 seeds with standard deviation.}
\end{table*}

\paragraph{Activation Quantization}
In this study, we explore the same range estimation methods for activation quantization. We keep the weights and backward pass in full precision and only quantize the activations using asymmetric uniform quantization. 

Similar to gradient quantization,  our method is on par with dynamic quantization methods and outperforms FP32. For activation quantization, in-hindsight min-max and running min-max perform significantly better than the other methods. DSGC performs worse in activation quantization, which is not surprising, as its objective function was designed with gradient distributions in mind. Once again current min-max trails the other methods.

\subsection{Fully Quantized Training Results}
\label{sec:results}
We now demonstrate the effectiveness of our method for fully quantized training on ImageNet and Tiny ImageNet benchmarks and compare against other dynamic quantization methods. We apply in-hindsight min-max to both activations and gradients. We also compare against using running  and current min-max for both tensors. The earlier is a variation of the method adopted by quantification parameter adjustment \cite{FX_backprop_trainig} but without the bit-with adjustment, whereas the latter is similar to the quantized training framework of \cite{chen2020statistical} when using per-tensor quantization. Finally, for Tiny ImageNet, we further compare to quantized training framework of DSGC without the adaptive learning rate.

\paragraph{Experimental Setup}
Weights, activations and gradients are quantized to 8-bits (W8/A8/G8). We use uniform stochastic quantization for gradients and standard uniform quantization for the weights and activations for all layers, including the first and last layer of the networks. The weights are always quantized with the current min-max method. For the DSGC method, we follow the author's approach and dynamically quantize activations using current min-max. 

We use a momentum of 0.9 for running min-max and in-hindsight min-max although we observe little sensitivity to that parameter. We also found that both methods benefit from an initial calibration step when used for activation quantization. By calibration, we mean feeding a few batches of data through the network to calibrate the quantization ranges before training starts.  

All models are trained using SGD with momentum of 0.9. ResNet18 is trained as described in the previous section. VGG16 is trained for 90 epochs with initial learning rate of 0.01. The learning rate is reduced by a factor of 10 at epochs 60 and 80. We train MobileNetV2 for 120 epochs with initial learning of 0.01 for the depthwise-separable layers and 0.1 for all other layers in the network. We found that this heterogeneous learning rate stabilizes quantized training for all methods. It also leads to no accuracy drop for FP32 training compared to a homogeneous learning rate of 0.1. We use a cosine annealing schedule with a final learning rate of $1\text{e-}5$ and a weight decay of $2\text{e-}5$.

\paragraph{Results Discussion}
% We observe that performs in line with the highest scoring dynamic method (running min-max) and outperforms all other methods on Tiny ImageNet. Current min-max and DSDG show 
% For ResNet-18 on ImageNet in-hindsight min-max peforms best. 

The results for Tiny ImageNet are shown in table \ref{tab:tiny_imagenet}. Across all 3 models, our hardware-friendly in-hindsight min-max performs on par with the significantly less efficient dynamic quantization methods. Only running min-max outperforms this slightly for MobileNetV2. 
We observe a similar trend for full ImageNet training (cf. table \ref{tab:imagenet}) where in-hindsight min-max performs similar to running min-max and marginally outperforms the commonly used current min-max.
In summary, for all cases, 8-bit in-hindsight quantization is close to the FP32 baseline (within .5\%) while fully utilizes the advantages of common fixed point accelerators.

\section{Memory Transfer Comparison}
\label{sec:memory_transfer_comparios}
\begin{figure*}
\begin{center}
\includegraphics[width=0.8\linewidth]{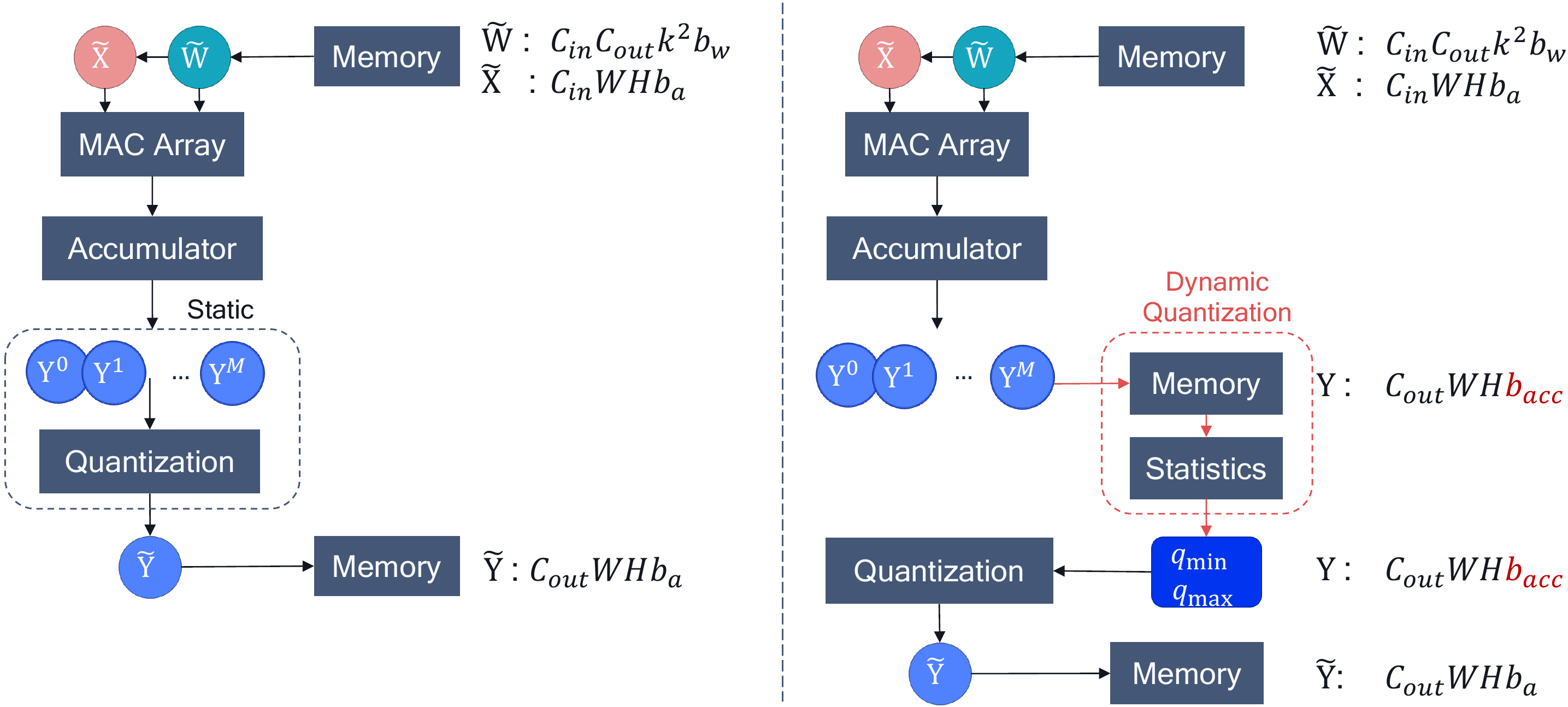}\vspace{-.5cm}
\end{center}
   \caption{Comparison of memory movements associated with static (left) and dynamic (right) quantization in a fixed-point neural network accelerator.}
\label{fig:memory_cost}
\end{figure*}

\begin{table*}
\begin{center}
\begin{tabular}{lccccccc}
\toprule
Network &Conv  & $C_{\text{in}}$  & $C_{\text{out}}$ & $W \times H$  & Static & Dynamic & Delta \\
\midrule
ResNet18 & 3$\times$3 & 64 & 64 & 56$\times$56    & 428 KB  & 1996 KB & +366\% \\ % block 1 Resnet18
ResNet18 & 3$\times$3 & 256 & 256 & 14$\times$14    & 674 KB  & 1066 KB & +58\% \\ % block 3 Resnet18
MobileNetV2 & 1$\times$1 & 16 & 96 & 112$\times$112    & 1374 KB  & 10782 KB & +685\% \\ % PW first inverted residual MNv2
MobileNetV2 & 3$\times$3 (DW) & 96 & 96 & 112$\times$112    & 882 KB  & 4410 KB & +400\% \\ % DW second inverted residual MNv2
MobileNetV2 & 3$\times$3 (DW) & 960 & 960 & 7$\times$7    & 100 KB  & 468 KB & +366\% \\ % DW close to last inverted residual MNv2
\bottomrule
\end{tabular}\vspace{-.4cm}
\end{center}
\caption{Memory movement costs comparison between static and dynamic quantization for various layers in ResNet18 and MobileNetV2 on ImageNet ($b_w=b_a=8$-bits, $b_{\text{acc}}=32$-bits, DW = depthwise separable).}
\label{tab:memory}
\end{table*}

%In this section we calculate the memory overhead associated with dynamic quantization for chosen layers from ResNet18 and MobileNetV2 during inference. 
In this section, we compare the memory transfer associated with static and dynamic quantization. We show it here for the forward pass, the backwards pass follows analogously (see figure \ref{fig:qt_framework}).

In static quantization, the quantized weight tensor $\widetilde{\mat{W}}$ ($C_{\text{in}}$ input channels, $C_{\text{out}}$ output channels, $k\times k$ kernel size) and the quantized input $\widetilde{\mat{X}}$ of size $W\times H$ are sequentially loaded to the MAC array. The output of the accumulator is then quantized and stored in memory. The total memory cost for static quantization in bits is given by:
\begin{equation}
 \label{eq:static_memory}
    \underbrace{C_{\text{in}}  C_{\text{out}}  k^2 b_w}_{\text{weight kernel}}
    + \underbrace{C_{\text{in}}  W  H b_a}_{\text{input feature}}
    + \underbrace{C_{\text{out}}  W  H b_a}_{\text{output feature}}
\end{equation}
where $b_a$ and $b_w$ are the activation and weight bit-width, respectively. 
As we discussed in section \ref{sec:dynamic_quantization},
dynamic quantization requires writing the accumulator output first in memory. After the statistics have been calculated the quantization parameters are updated and the output is then brought back to the compute unit to be quantized. The quantized output is then written back to memory. To total memory cost of dynamic quantization is given by:
\begin{equation}
\label{eq:dynamic_memory}
\begin{split}
    \underbrace{C_{\text{in}} C_{\text{out}} k^2 b_w}_{\text{weight kernel}}
    + \underbrace{C_{\text{in}} W H b_a}_{\text{input feature}}
    + \underbrace{C_{\text{out}} W H b_{\text{acc}}}_{\text{save acc output}}\\
    + \underbrace{C_{\text{out}} W H b_{\text{acc}}}_{\text{load acc output}}
    + \underbrace{C_{\text{out}} W H b_{a}}_{\text{save quantized output}}
\end{split}
\end{equation}
where $b_{\text{acc}}$ is the bit-width of the accumulator. 
Figure \ref{fig:memory_cost} illustrates the memory movement associated with every step.

In table \ref{tab:memory}, we compare the memory movement cost of static and dynamic quantization for typical layers in ResNet18 and MobileNetV2 using $b_w=b_a=8$-bits and $b_{\text{acc}}=32$-bits.
The exact overhead of dynamic quantization depends on many parameters, such as the input size, number of channels and type of kernel. In most cases, the extra memory movement is about $4\times$. Only in later layers in ResNet18, where the weight tensor is significantly larger than the input feature map, is the overhead lower. In the extreme case of certain point-wise convolutions in MobileNetV2, the memory movement of dynamic quantization can be  $8\times$ higher than for static quantization.

\section{Conclusions}
\label{sec:conclusions}
In this paper, we provide a general framework for estimating quantization ranges in the context of quantized training that overcomes the need for dynamic quantization. Our proposed approach, \textit{in-hindsight range estimation}, uses past estimates of the quantization parameters to enable static quantization of the tensor in question. It relies on extracting simple statistics from the output tensor in an online fashion to update the quantization parameters for the next iteration.  These statistics need to be extracted from the accumulator before the quantization step, which may require appropriate hardware logic. While this is a general framework, we propose a specific variant, called \textit{in-hindsight min-max} that uses the min-max statistics. 

We demonstrate the effectiveness of our methods on popular image classification benchmarks by comparing them to other dynamic quantization techniques found in literature. We show that \textit{in-hindsight min-max} performs on par with the best scoring dynamic range methods while reducing significantly the memory overhead associated with dynamic quantization. It can be used as a drop-in replacement method for estimating quantization ranges that can better utilize the common fixed-point accelerators for quantized training.

{\small
\bibliographystyle{ieee_fullname}
\bibliography{egbib}
}

\end{document}